\documentclass{article}

\PassOptionsToPackage{numbers, compress}{natbib}
\usepackage{multirow}
\usepackage{caption}
\usepackage{graphicx}
\usepackage{pgfplots}
\pgfplotsset{compat=1.18}

\usepackage[final]{d3s3_neurips_2024}
\usepackage[utf8]{inputenc} % allow utf-8 input
\usepackage[T1]{fontenc}    % use 8-bit T1 fonts
\usepackage{hyperref}       % hyperlinks
\usepackage{url}            % simple URL typesetting
\usepackage{booktabs}       % professional-quality tables
\usepackage{amsfonts}       % blackboard math symbols
\usepackage{nicefrac}       % compact symbols for 1/2, etc.
\usepackage{microtype}      % microtypography
\usepackage{xcolor}         % colors
\usepackage{soul}

\title{Guaranteeing Conservation Laws with Projection in Physics-Informed Neural Networks}

\author{%
  Anthony Baez \\
  MIT\\
  \And
  Wang Zhang \\
  MIT\\
  \And  
  Ziwen Ma \\
  MIT\\
  \And
  Subhro Das \\
  IBM Research\\
  \And
  Lam M. Nguyen \\
  IBM Research\\
  \And
  Luca Daniel \\
  MIT\\
}

\begin{document}

\maketitle

\begin{abstract}

Physics-informed neural networks (PINNs) incorporate physical laws into their training to efficiently solve partial differential equations (PDEs) with minimal data \cite{raissi2019physics}. However, PINNs fail to guarantee adherence to conservation laws, which are also important to consider in modeling physical systems. To address this, we proposed PINN-Proj, a PINN-based model that uses a novel projection method to enforce conservation laws. We found that PINN-Proj substantially outperformed PINN in conserving momentum and lowered prediction error by three to four orders of magnitude from the best benchmark tested. PINN-Proj also performed marginally better in the separate task of state prediction on three PDE datasets.

\end{abstract}

\section{Introduction}

% Explain what PINNs are, why they are useful

Physics-informed neural networks (PINNs) \cite{raissi2019physics} are a class of neural networks that incorporate laws of physics, e.g. partial differential equations (PDEs), to efficiently and accurately learn to solve PDEs from a small sample of data. In PINNs, the physics equation is assumed to be partially known and is incorporated into the loss function of the neural network. This creates a soft constraint where the PINN is penalized for producing outputs that violate the physical equation. This approach allows flexibility during training so the PINN can balance both fitting the observed data and conforming to the governing PDE.

This approach, however, permits PINN to violate conservation laws, which are fundamental principles of physics. A PINN that always follows conservation laws would be more accurate and trustworthy when applied to model real-world systems. Adding another soft constraint would not address this issue. Instead, we could impose a hard constraint that ensures the PINN's prediction always obeys conservation regardless of the nature of the data. 
 
In this work, we propose a novel projection technique that can guarantee adherence to conservation laws in a PINN by projecting its output into a functional space where the conservation law is not violated. This projection approach establishes a rigid, inviolable boundary that also allows adaptability during training. The projection is applied at both training and testing, so the model learns to produce outputs that observe conservation laws and accurately represent the PDE.

To evaluate the effectiveness of this projection method, we compared a PINN model modified with the projection method, referred to as PINN-Proj, to a vanilla PINN and a PINN model with a soft constraint on the conservation law, referred to as PINN-SC. These models were trained and evaluated on three physics PDEs: the Advection equation, the viscous Burgers' equation, and the Korteweg-De Vries equation. The models were evaluated on the error of its predictions of the state $u$ and the conserved quantity $c$. We found that PINN-Proj significantly outperformed PINN on the conservation 
task while performing marginally better than PINN on the state prediction task.

\section{Related Work}

In previous work, the conservation law has been incorporated into the PINN by adding another term to the loss function, which creates a soft constraint. Training can either be done entirely with this modified loss function \cite{wu2022prediction}, or in two stages where the modified loss function is used in the second stage of training \cite{lin2022two}. The physics PDE can also be completely replaced by a Hamiltonian \cite{greydanus2019hamiltonian} or a Lagrangian \cite{cranmer2019lagrangian}, which describe how the energy of a physical system is conserved.

Another branch of previous work uses hard constraints in PINNs. In topology optimization, a hard constraint can be used to keep the volume of the fluid described by the PDE below a certain value \cite{lu2021physics}. The boundary conditions of the physics PDE can be incorporated as hard constraints as well, and the addition of this constraint has been found to lower testing error \cite{negiar2023learning}. A projection incorporated as a layer in a neural network has also been used as a hard constraint to guarantee that a non-PINN neural network preserves learned conservation laws in a dynamical system \cite{zhang2023concernet}.

\section{Method}

\subsection{Physics-informed Neural Network}

The physics-informed neural network (PINN) can be defined as 

\begin{equation}
f = u_{t} + \mathcal{N}[u]
\end{equation}

where $u(x, t)$ is the solution to the equation and a function of spatial coordinate $x$ and time coordinate $t$. $u(x, t)$ is parameterized by a neural network. $\mathcal{N[\cdot]}$ is a differentiable nonlinear operator that acts on $u(x, t)$. The PINN can learn a solution for the governing PDE using the loss function, $\mathcal{L}$, defined as

\begin{equation}
\mathcal{L} = \frac{1}{N_{u}} \sum_{i=1}^{N_{u}} {|u(x^{i}_{u}, t^{i}_{u}) - u^{i}|^{2}} + \frac{1}{N_{f}} \sum_{i=1}^{N_{f}} {|f(x^{i}_{f}, t^{i}_{f})|^{2}}
\end{equation}

Here, $\{{x^{i}_{u}, t^{i}_{u}, u^{i}}\}$ are the initial and boundary data on $u(x, t)$, while $\{{x^{i}_{f}}, t^{i}_{f}\}$ are collocation points. The first term of $\mathcal{L}$ is the mean squared error between the PINN prediction $u(x, t)$ and ground truth $u^{i}$. The second term of $\mathcal{L}$ is the mean squared error of $f(x, t)$, as $f$ would be zero if a perfect solution was found. We will refer to $u(x, t)$ as the state of the PDE.

\subsection{Conservation Law}

Let us begin with momentum as our conserved quantity. The physical systems we used with PINN are subject to the law of conservation of momentum, as they do not have inflow or outflow of momentum. If we choose a PDE where the state, $u$, is the velocity of a fluid, then the total momentum of the system, which we define as our conserved quantity $c$, at time $t$ would be

\begin{equation}
c(t) = \int_{X} u(x, t) \, dx \approx \sum_{x \in X} {u(x, t)} \Delta x 
\end{equation}

where $X$ is the spatial domain. If momentum is conserved in the system, $c(t)$ would remain constant for all times $t$.

\subsection{Projection Method}

The prediction of the state from PINN-Proj, $u_{proj}(x, t)$, is defined as

\begin{equation}
u_{proj}(x, t) = u(x, t) - \int_{X}\frac{u(x', t)}{|X|}  dx' + \frac{c}{|X|}
\end{equation}

where $u(x, t)$ is the nominal prediction of the state from PINN. In our projection equation, the second term calculates and removes the momentum of the system at time $t$ using a dummy variable $x'$, and the third term adds the known conserved momentum value to the system. This projects the output of the neural network into a space where conservation of momentum is guaranteed. Now using $u_{proj}(x, t)$ as the prediction for the state, we use Equation 3 to calculate the total momentum of the system predicted by PINN-Proj,

\begin{equation}
\int_{X} u_{proj}(x, t) \, dx = \int_{X} u(x, t) dx - \int_{X} \int_{X} \frac{u(x', t) }{|X|} dx' dx + \int_{X} \frac{c}{|X|} dx
\end{equation}

\begin{equation}
= \int_{X} u(x, t) dx - \int_{X} u(x', t) dx' + c = c
\end{equation}

which is always $c$ and is therefore always conserved.

\subsection{Experimental Setup}

We trained three different models: PINN, PINN-SC, PINN-Proj. A soft constraint was added to PINN-SC by adding the mean squared error between the current momentum (second term of Equation 4) and the conserved momentum (third term of Equation 4) to the loss function. A coefficient of 10 was multiplied to this term as well. The training setup is the same as the original PINN method \cite{raissi2019physics} unless otherwise mentioned.  The models were trained on three datasets which contained $x$, $t$, and $u$ values generated from the Advection Equation \cite{takamoto2022pdebench}, the viscous Burgers' Equation \cite{raissi2019physics}, and the Korteweg-De Vries Equation \cite{raissi2019physics}. Each PDE dataset contained 256 $x$ values, 100 $t$ values, and 25,600 total state values $u$. All PDE datasets modeled conserved systems, so there was no input or output of momentum into the systems. 

The training set contained 100 points that were randomly sampled from each PDE dataset and 10,000 collocation points calculated using Latin Hypercube Sampling. Each trial had a different random sample from its PDE dataset \footnote{All code is available at \url{https://github.com/antbaez9/pinn-proj}}. More details about the setup can be found in the Appendix.

\subsection{Evaluation}

The accuracy of PINN, PINN-SC, and PINN-Proj was measured on the metric of predicting state values, $u$, and on conservation of the conserved quantity, $c$. The errors of these tasks were referred to as Error $u$ and Error $c$, respectively. The ground truth of $c$ was the mean total momentum over all $t$ values for each dataset. Relative $\mathcal{L}_2$ error was used for Error $u$, and absolute $\mathcal{L}_2$ error was used for Error $c$ because the ground truth total momentum of all three systems is zero. The average errors across five trials were reported.

\section{Results}

\begin{table}[h]
\centering
\renewcommand{\arraystretch}{1.4}
\begin{tabular}{*{7}{c}}
\hline
\multirow{2}{*}{PDE} & \multicolumn{2}{c}{PINN} & \multicolumn{2}{c}{PINN-SC} & \multicolumn{2}{c}{PINN-Proj} \\
\cline{2-7}
 & Error $u$ & Error $c$ & Error $u$ & Error $c$ & Error $u$ & Error $c$ \\
\hline
Advection Eq. & \textbf{2.10E-03} & 3.85E-02 & 5.13E-03 & 4.17E-03 & 2.24E-03 & \textbf{1.31E-06} \\
\hline
Burgers' Eq. & 2.80E-03 & 3.02E-02 & 2.16E-03 & 2.12E-03 & \textbf{1.82E-03} & \textbf{1.58E-06} \\
\hline
Korteweg-De Vries Eq. & 3.02E-02 & 5.55E-01 & 2.34E-02 & 2.40E-02 & \textbf{1.65E-02} & \textbf{1.65E-06} \\
\hline
\end{tabular}
\vspace{5pt}
\caption{Average $\mathcal{L}_2$ error for state prediction task (Error $u$) and conservation task (Error $c$)}
\label{tab:example}
\end{table}

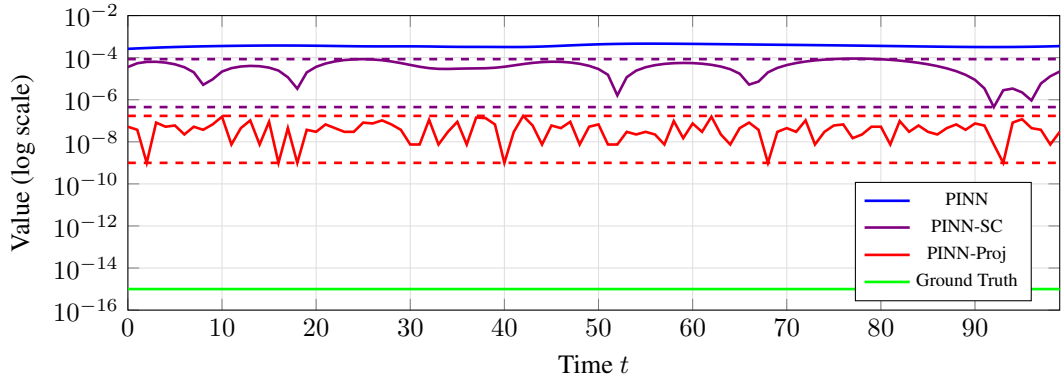
\begin{figure}
    \centering
    \begin{tikzpicture}
        \begin{semilogyaxis}[
            width=\textwidth,
            height=5.5cm,
            xlabel={Time $t$},
            ylabel={Value (log scale)},
            xmin=0, xmax=99,
            ymin=1e-16, ymax=1e-2,  % Adjusted for log scale
            xtick distance=10,
            ytick={1e-16,1e-14,1e-12,1e-10,1e-8,1e-6,1e-4,1e-2},
            yticklabels={$10^{-16}$,$10^{-14}$,$10^{-12}$,$10^{-10}$,$10^{-8}$,$10^{-6}$,$10^{-4}$,$10^{-2}$,$1$},
            grid=both,
            minor grid style={gray!25},
            major grid style={gray!25},
            legend pos=south east,
        legend style={
            font=\scriptsize,  % Adjust font size of legend text
            },
        ]

        % Original data (blue line)
        \addplot[
            blue,
            mark=none,
            line width=1pt
            ] 
            coordinates {
            (0,0.00026268) (1,0.00027555) (2,0.00028795) (3,0.00029975) (4,0.00031078)
            (5,0.00032079) (6,0.00032985) (7,0.00033814) (8,0.00034559) (9,0.0003522)
            (10,0.0003584) (11,0.00036365) (12,0.00036818) (13,0.00037211) (14,0.00037503)
            (15,0.00037694) (16,0.00037789) (17,0.00037736) (18,0.00037575) (19,0.00037289)
            (20,0.00036895) (21,0.00036418) (22,0.00035888) (23,0.00035363) (24,0.0003491)
            (25,0.00034559) (26,0.00034386) (27,0.0003435) (28,0.00034428) (29,0.00034535)
            (30,0.00034547) (31,0.00034368) (32,0.00033975) (33,0.00033456) (34,0.00032949)
            (35,0.00032622) (36,0.00032479) (37,0.00032473) (38,0.00032437) (39,0.00032306)
            (40,0.00032121) (41,0.00032032) (42,0.00032228) (43,0.00032789) (44,0.00033754)
            (45,0.00035077) (46,0.00036645) (47,0.00038332) (48,0.00040025) (49,0.0004161)
            (50,0.00043017) (51,0.00044191) (52,0.00045079) (53,0.00045735) (54,0.0004614)
            (55,0.00046325) (56,0.00046337) (57,0.00046182) (58,0.00045913) (59,0.00045556)
            (60,0.00045133) (61,0.00044686) (62,0.00044203) (63,0.00043726) (64,0.00043243)
            (65,0.00042781) (66,0.00042325) (67,0.00041893) (68,0.00041461) (69,0.00041053)
            (70,0.00040647) (71,0.00040242) (72,0.00039837) (73,0.00039417) (74,0.00038996)
            (75,0.00038576) (76,0.00038126) (77,0.00037673) (78,0.00037199) (79,0.00036711)
            (80,0.00036216) (81,0.00035718) (82,0.0003522) (83,0.00034732) (84,0.00034246)
            (85,0.00033787) (86,0.00033349) (87,0.00032967) (88,0.00032622) (89,0.00032341)
            (90,0.00032121) (91,0.00031987) (92,0.00031963) (93,0.00032032) (94,0.00032225)
            (95,0.00032553) (96,0.0003303) (97,0.00033668) (98,0.00034478) (99,0.00035468)
            };
        \addlegendentry{PINN}

        \addplot[
            violet,
            mark=none,
            line width=1pt,
            ] 
            coordinates {
            (0,3.5762787e-05) (1,5.4240227e-05) (2,6.300211e-05) (3,6.3717365e-05) (4,5.8293343e-05)
            (5,4.8339367e-05) (6,3.5107136e-05) (7,2.0205975e-05) (8,5.1259995e-06) (9,9.2983246e-06)
            (10,2.1874905e-05) (11,3.1709671e-05) (12,3.8027763e-05) (13,4.0650368e-05) (14,3.9219856e-05)
            (15,3.361702e-05) (16,2.43783e-05) (17,1.168251e-05) (18,3.3378601e-06) (19,1.9907951e-05)
            (20,3.6716461e-05) (21,5.2630901e-05) (22,6.6399574e-05) (23,7.6949596e-05) (24,8.3267689e-05)
            (25,8.5115433e-05) (26,8.225441e-05) (27,7.5519085e-05) (28,6.6041946e-05) (29,5.531311e-05)
            (30,4.5180321e-05) (31,3.695488e-05) (32,3.1769276e-05) (33,2.9325485e-05) (34,2.9087067e-05)
            (35,2.9861927e-05) (36,3.0577183e-05) (37,3.0934811e-05) (38,3.1411648e-05) (39,3.3140182e-05)
            (40,3.695488e-05) (41,4.273653e-05) (42,4.9710274e-05) (43,5.6445599e-05) (44,6.1631203e-05)
            (45,6.4194202e-05) (46,6.3300133e-05) (47,5.8948994e-05) (48,5.120039e-05) (49,4.0948391e-05)
            (50,2.8848648e-05) (51,1.5377998e-05) (52,1.6093254e-06) (53,1.180172e-05) (54,2.4199486e-05)
            (55,3.5047531e-05) (56,4.3988228e-05) (57,5.0604343e-05) (58,5.4657459e-05) (59,5.6147575e-05)
            (60,5.5193901e-05) (61,5.1617622e-05) (62,4.5716763e-05) (63,3.7819147e-05) (64,2.8252602e-05)
            (65,1.7255545e-05) (66,5.2452087e-06) (67,7.301569e-06) (68,2.0056963e-05) (69,3.2633543e-05)
            (70,4.452467e-05) (71,5.5581331e-05) (72,6.5505505e-05) (73,7.3999166e-05) (74,8.0734491e-05)
            (75,8.5890293e-05) (76,8.9079142e-05) (77,9.0539455e-05) (78,9.0092421e-05) (79,8.7887049e-05)
            (80,8.4191561e-05) (81,7.891655e-05) (82,7.2538853e-05) (83,6.5028667e-05) (84,5.6922436e-05)
            (85,4.8398972e-05) (86,3.9637089e-05) (87,3.0934811e-05) (88,2.2828579e-05) (89,1.5258789e-05)
            (90,8.8214874e-06) (91,3.4570694e-06) (92,4.4703484e-07) (93,2.8312206e-06) (94,3.3974648e-06)
            (95,2.2351742e-06) (96,9.2387199e-07) (97,5.9008598e-06) (98,1.3053417e-05) (99,2.1934509e-05)
            };
        \addlegendentry{PINN-SC}

        % New data set (red line)
        \addplot[
            red,
            mark=none,
            line width=1pt,
            ] 
            coordinates {
            (0,5.2154064e-08) (1,3.7252903e-08) (2,1e-9) (3,8.1956387e-08) (4,5.2154064e-08)
            (5,5.9604645e-08) (6,2.2351742e-08) (7,5.2154064e-08) (8,3.7252903e-08) (9,6.7055225e-08)
            (10,1.5646219e-07) (11,7.4505806e-09) (12,4.4703484e-08) (13,1.0430813e-07) (14,7.4505806e-09)
            (15,8.9406967e-08) (16,1e-9) (17,4.4703484e-08) (18,1e-9) (19,3.7252903e-08)
            (20,2.9802322e-08) (21,6.7055225e-08) (22,4.4703484e-08) (23,2.9802322e-08) (24,2.9802322e-08)
            (25,8.1956387e-08) (26,7.4505806e-08) (27,1.0430813e-07) (28,6.7055225e-08) (29,3.7252903e-08)
            (30,7.4505806e-09) (31,7.4505806e-09) (32,1.1175871e-07) (33,7.4505806e-09) (34,2.9802322e-08)
            (35,8.9406967e-08) (36,7.4505806e-09) (37,1.4156103e-07) (38,1.3411045e-07) (39,6.7055225e-08)
            (40,1e-9) (41,2.9802322e-08) (42,1.7136335e-07) (43,5.9604645e-08) (44,7.4505806e-09)
            (45,5.9604645e-08) (46,4.4703484e-08) (47,8.9406967e-08) (48,7.4505806e-09) (49,5.2154064e-08)
            (50,6.7055225e-08) (51,7.4505806e-09) (52,7.4505806e-09) (53,2.9802322e-08) (54,2.2351742e-08)
            (55,2.9802322e-08) (56,2.2351742e-08) (57,7.4505806e-09) (58,9.6857548e-08) (59,1.4901161e-08)
            (60,7.4505806e-08) (61,2.9802322e-08) (62,1.5646219e-07) (63,2.2351742e-08) (64,7.4505806e-09)
            (65,4.4703484e-08) (66,8.1956387e-08) (67,3.7252903e-08) (68,1e-9) (69,2.9802322e-08)
            (70,4.4703484e-08) (71,2.9802322e-08) (72,1.0430813e-07) (73,7.4505806e-09) (74,2.9802322e-08)
            (75,5.9604645e-08) (76,6.7055225e-08) (77,1.4901161e-08) (78,2.2351742e-08) (79,5.2154064e-08)
            (80,5.2154064e-08) (81,7.4505806e-09) (82,9.6857548e-08) (83,2.9802322e-08) (84,5.9604645e-08)
            (85,2.9802322e-08) (86,2.2351742e-08) (87,4.4703484e-08) (88,6.7055225e-08) (89,3.7252903e-08)
            (90,7.4505806e-08) (91,5.9604645e-08) (92,7.4505806e-09) (93,1e-9) (94,8.1956387e-08)
            (95,1.1920929e-07) (96,4.4703484e-08) (97,3.7252903e-08) (98,7.4505806e-09) (99,2.9802322e-08)
            };
        \addlegendentry{PINN-Proj}

        % Horizontal line at y=1e-6 (as a reference line)
        \addplot[
            green,
            line width=1pt,  % Made this line slightly thicker for emphasis
            ] 
            coordinates {
            (0,1e-15) (99,1e-15)
            };
        \addlegendentry{Ground Truth}

        \addplot[
            violet,
            line width=1pt, 
            dashed,
            opacity=0.5
            ] 
            coordinates {
            (0,8.5115433e-05) (99,8.5115433e-05)
            };
        \addplot[
            violet,
            line width=1pt, 
            dashed,
            opacity=0.5
            ] 
            coordinates {
            (0,4.4703484e-07) (99,4.4703484e-07)
            };
        \addplot[
            red,
            line width=1pt, 
            dashed,
            opacity=0.5
            ] 
            coordinates {
            (0,1e-9) (99,1e-9)
            };
        \addplot[
            red,
            line width=1pt, 
            dashed,
            opacity=0.5
            ] 
            coordinates {
            (0,1.71E-07) (99,1.71E-07)
            };

        \end{semilogyaxis}
    \end{tikzpicture}
    \caption{Predicted and ground truth of $c$ over time for first trial of Burgers' Equation }
    \label{fig:ct_pred_comparison_logscale}
\end{figure}

Table 1 shows the results of training PINN, PINN-SC, and PINN-Proj on our PDE datasets. On the task of predicting the state, PINN-Proj has the lowest Error $u$ on the Burgers' and Korteweg-De Vries Equation datasets, scoring 3.4E-04 and 6.9E-03 lower than that of the next best model on that task, PINN-SC, respectively. PINN has the lowest Error $u$ on the Advection dataset, scoring 1.4E-04 lower than that of the next best model on that task, PINN-Proj.

On the task of maintaining the conserved quantity, PINN-Proj significantly outperforms PINN and PINN-SC on all three datasets. On the Advection Equation and Burgers' Equation dataset, PINN-Proj outperforms the next best model, PINN-SC, by three orders of magnitude, and on the Korteweg-De Vries Equation dataset, PINN-Proj outperforms PINN-SC by four orders of magnitude.

We can see in Figure 1 the total momentum of PINN, PINN-SC, and PINN-Proj at each time $t$, which are their predictions of the conserved quantity $c$. The $c$ values in Figure 1 also reflect the results in Table 1, as PINN-Proj is more accurate than PINN-SC, which is in turn more accurate than PINN. The dashed lines show the upper and lower bounds for the values in PINN-SC and PINN-Proj.

\section{Discussion}

On Error $u$, PINN-Proj performs best on two datasets and PINN performs best on one dataset. Both margins that PINN-Proj performs better than the next best model are higher than the PINN's margin of outperformance. The results, therefore, demonstrate that the projection method is able to reliably and significantly improve the conservation of the conserved quantity, and even marginally improve performance on the state prediction task compared to using a soft constraint or no constraint in PINN.

Across all three datasets, Error $c$ for the PINN-Proj method is near 1E-6. One possible cause for the error not being lower could be the discretization of the $x$ and $t$ domains of all datasets was done to five digits, which is just under the order of magnitude of the Error $c$. Another pattern in the $c$ values is that PINN-Proj, and to a lesser extent PINN-SC, have more variation than in PINN. This variation is likely due to the many floating-point calculations during the integration in PINN-SC and PINN-Proj and is lower in PINN-SC because of the flexibility of the soft constraint. This could also be worsened by the discretization accuracy of $x$ and $t$. The projection method significantly increased training time as well, since the integration involved many calculations at every epoch during training. These issues of accuracy, variability, and computational cost will be further addressed in future work.

% The ground truth $c$ value of around 1E-15 is likely due to floating-point error and should be 0. However, since the predicted $c$ values for all our PINN models are significantly above this value and the same ground truth values were applied to all models, floating-point error is likely not a significant source of error in the calculation of Error $c$.

% In Figure 1, the predicted $c$ value for PINN-Proj is notably more variable than the other PINN models. This could be inherent to the projection layer, specifically the integration which is done with discrete values. This could also be worsened by the dicretization accuracy also noted. More investigation is needed. 

The main limitation of our method is that it is only applicable to systems with no net input or output to the system over time, which is a special case of the Neumann boundary condition. So, future work will involve extending the projection method to general cases of different boundary conditions. We also were limited by solvers available to generate our own dataset, so we relied on previously generated datasets and benchmarks. If we were to use a dedicated solver, it would allow more control, consistency, and a wider range of PDEs to be tested. 

\section{Conclusion}

This paper proposes a novel projection method to guarantee conservation in a PINN. To determine its efficacy, we compared a PINN that uses the projection method, PINN-Proj, to an unmodified PINN and a PINN with a soft constraint, PINN-SC. After applying these models to datasets generated from the Advection, Burgers', and Korteweg-De Vries PDEs, we found that PINN-Proj, while adding computational cost, greatly improves and guarantees conservation of momentum while marginally improving performance on state prediction in conserved systems when compared to PINN. 

\bibliographystyle{plain}  % Choose the bibliography style you prefer
\bibliography{d3s3_neurips_2024}  % This links to your .bib file named references.bib

%%%%%%%%%%%%%%%%%%%%%%%%%%%%%%%%%%%%%%%%%%%%%%%%%%%%%%%%%%%%

\appendix

\section{Appendix}

The PINN models contained 9 hidden linear layers of 20 neurons. The weights of the neural network were initialized using Xavier normal initialization. All activation functions were hyperbolic tangent. The optimizer was L-BFGS. The L-BFGS optimizer used stops itself when optimality conditions are met, so the number of epochs of training varied between trials and between datasets.

The Advection equation had a velocity of flow of 0.1 and its dataset was generated using PDEBench. The viscous Burgers' equation had a diffusion constant of 0.1. The Korteweg-De Vries equation had a dispersive term of 0.0025.

%%%%%%%%%%%%%%%%%%%%%%%%%%%%%%%%%%%%%%%%%%%%%%%%%%%%%%%%%%%%

\end{document}